\documentclass[11pt]{article}
\usepackage[a4paper,margin=23mm]{geometry}
\usepackage[T1]{fontenc}
\usepackage{lmodern,microtype}
\usepackage{amsmath,amssymb,graphicx,booktabs,tabularx,array,multirow}
\usepackage{xcolor,caption,subcaption,float,placeins}
\usepackage[numbers,sort&compress]{natbib}
\usepackage[breaklinks,colorlinks=true,linkcolor=blue!45!black,citecolor=blue!45!black,urlcolor=blue!45!black]{hyperref}
\usepackage{xurl}
\usepackage{fancyhdr}
\hypersetup{pdftitle={PAANI: On-Device Visual Evidence Fusion and Explainable Guidance for River-Robot Simulation},pdfauthor={Savio Cardoz and Santhiya Rajan},pdfsubject={On-device AI and robotics systems architecture and evaluation}}
\newcolumntype{Y}{>{\raggedright\arraybackslash}X}

\newcommand{\clip}{\operatorname{clip}}

\newcommand{\E}{\mathcal{E}}
\title{\textbf{PAANI: On-Device Visual Evidence Fusion and Explainable Guidance for River-Robot Simulation}}
\author{Savio Cardoz$^{1}$ \qquad Santhiya Rajan$^{2,3}$\thanks{This project was undertaken by both authors in their personal capacities, outside their professional employment. Affiliations are provided for identification only and do not imply institutional involvement, sponsorship or endorsement.}\\[0.7em]
\small $^{1}$ACL Digital, Pune\\
\small $^{2}$Multiverse Computing, Spain\\
\small $^{3}$Ph.D. (Part-time) Research Scholar,\\
\small PSG College of Technology, Coimbatore, India}
\date{}
\begin{document}
\maketitle
\begin{abstract}
Mobile river-monitoring robots must interpret obstacles and water boundaries that geographic waypoints alone cannot describe. On resource-constrained platforms, converting imperfect visual predictions into timely, inspectable guidance is a distinct challenge: an object label or steering command does not explain which evidence supports a decision or when that evidence is unreliable. We present PAANI, an on-device perception-to-guidance architecture that combines a project-fine-tuned YOLO11n detector and a custom MobileNetV3-Small semantic segmenter with timestamp-aligned evidence fusion on Arduino UNO Q. Bounded tracking supplies object persistence, while an explicit corridor policy blends surface labels, accepted detections, legacy urgency and mask uncertainty. Each final advisory exposes its contributing evidence and policy reasons. ROS~2 interfaces connect this local AI pipeline to a separate Gazebo vessel, localization and control testbed. Training uses 10,000 WaterScenes images for four-class detection and 1,127 MaSTr1325 images for segmentation, with 198 segmentation validation images. The selected FP32 ONNX pair occupies 14.817~MB. Detector checkpoint test mAP@0.5 is 0.7388; the separately evaluated rectangular ONNX export reaches validation mAP@0.5 of 0.7367. Segmentation ONNX validation mIoU is 0.9750. A five-minute UNO Q recording yields median/p95 pipeline latency of 467.8/580.3~ms at a configured 0.5~Hz cadence. The evaluation also exposes black-input misclassification and a sampling-rate mismatch that prevents the diagnostic apparent-motion estimator from collecting sufficient evidence. These results support an inspectable, reusable edge-robotics foundation, while distinguishing model accuracy and on-board execution from validated on-water collision avoidance.
\end{abstract}

\section{Introduction}
Mobile river observations are useful only when the platform carrying the instruments can operate within a defensible sensing and control envelope. Reliable guidance, navigation, collision avoidance and fault handling remain recurring barriers to higher USV autonomy \cite{campbell2012,liu2023survey}. A geographic waypoint identifies a desired location, but says little about a vessel crossing the view, a pier extending into a channel, or a reflection obscuring the water boundary. The robotics problem is consequently not solved by attaching a camera to a route follower. It requires an interface between visual interpretation, uncertainty, temporal consistency and downstream control.

This work addresses that interface. A generic detector supplies object labels but not a complete representation of the water surface. A color threshold supplies surface evidence but is sensitive to appearance. A single-frame decision can flicker, while a steering arrow without provenance provides little help when diagnosing a mistake. These observations motivate complementary learned perception, bounded history, and explanations tied to the implemented policy rather than a post-hoc narrative detached from it.

PAANI combines two project-trained models with explicit tracking and advisory logic on the Linux side of Arduino UNO Q. The microcontroller provides indication and a communication watchdog, while an external host runs the simulator. The current demonstration uses river imagery that is independent of the simulated vehicle pose. This is a useful integration boundary: it permits exercising the board-side perception and ROS interfaces, but it does not close the visual feedback loop through scene geometry.

The contribution is an engineering research prototype, not a new foundational neural architecture. Specifically, we present: (i) a compact, versioned two-model RGB pipeline adapted to river-navigation labels; (ii) frame-aligned evidence fusion with inspectable risk components, temporal acceptance and final-decision explanations; (iii) a split edge/simulation robotics architecture that separates advice from actuator interfaces; and (iv) a documented analysis of model fitting, export quality, operational timing, diagnostic behavior and known failures. The central thesis is that reusable local robotics intelligence should expose evidence that developers can understand, test and improve.

\section{Related work and positioning}
WaterScenes provides multimodal water-surface perception data and multiple annotation tasks, including object and free-space interpretation \cite{waterscenes}. PAANI uses its RGB images and detection labels, not its radar modality; it therefore cannot inherit radar-derived range or velocity claims. MaSTr1325 targets maritime semantic segmentation and contains coastal USV imagery with water, sky and obstacle/environment labels \cite{mastr}. LaRS extends maritime evaluation to diverse lakes, rivers and seas, panoptic labels, scene attributes and temporal context \cite{lars}. SeaDronesSee instead addresses aerial detection and tracking over open water \cite{seadronessee}; its viewpoint illustrates why a maritime label set alone does not make datasets interchangeable. PAANI's completed training uses WaterScenes and MaSTr1325, so neither the broader LaRS coverage nor aerial-view results are claimed.

MODS emphasizes obstacle-oriented evaluation rather than treating high aggregate pixel accuracy as sufficient for USV navigation \cite{mods}. WaSR combines visual and inertial cues for water-obstacle separation \cite{wasr}; weak-annotation scaffolding addresses the cost of dense maritime masks \cite{scaffolding}; and WaSR-T uses temporal appearance to suppress reflection and glitter false positives \cite{wasrt}. Embedded maritime networks such as eWaSR address deployment-oriented perception \cite{ewasr}, while the MaCVi challenges explicitly evaluate both obstacle quality and embedded operation \cite{macvi2024}. These works motivate boundary-sensitive, deployment-aware evaluation, but PAANI does not reproduce their architectures or report a shared-benchmark comparison.

PAANI's selected segmenter is a custom lightweight MobileNetV3-Small encoder with an FPN-style decoder. MobileNetV3 provides an efficient pretrained representation \cite{mobilenet}, top-down feature-pyramid fusion supplies multi-scale structure \cite{fpn}, and encoder-decoder segmentation literature highlights the importance of recovering boundary detail \cite{deeplabv3plus}. The detector uses the Ultralytics YOLO11 nano implementation \cite{yolo11}. Its lightweight tracker is also deliberately narrower than established tracking-by-detection systems such as SORT \cite{sort}: it uses greedy IoU association and bounded history, without a motion filter or learned re-identification. These choices favor inspectability and bounded CPU work, not state-of-the-art tracking accuracy.

Confidence and explanation require equally careful scope. Modern neural-network confidence can be miscalibrated \cite{guo2017}, and high softmax scores do not by themselves solve misclassification or distribution-shift detection \cite{hendrycks2017}. Explainable-robotics literature distinguishes the many audiences and forms of a robot explanation \cite{sakai2022}. Model-agnostic methods such as LIME explain a prediction through a local surrogate \cite{ribeiro2016}, whereas critiques of post-hoc explanation emphasize that an explanation should not be treated as a substitute for an interpretable or validated decision process \cite{rudin2019}. PAANI takes a narrower systems approach: it exposes the evidence and deterministic policy path after the learned perception stages. This is a faithful trace of the implemented fusion logic, not a causal interpretation of neural features and not a guarantee that the source predictions are correct.

ROS~2 supplies modular communication and common robotics interfaces \cite{ros2}. PAANI uses these interfaces to isolate perception, navigation and control, rather than to claim hard real-time guarantees. Gazebo has a long history as an open robotics simulation environment \cite{koenig2004}, and generalized ROS EKF implementations provide a standard pattern for heterogeneous state estimation \cite{moore2014}. In PAANI, however, simulation fidelity and localization availability do not close the independent camera-to-world loop. The cited platforms establish engineering context, not evidence of field autonomy.

\section{System architecture and design}
\subsection{Compute partition and operational scope}
Unless explicitly marked otherwise, the architecture describes the recorded trained-model deployment. Model selection, thread settings and simulation configuration are reported for the evaluated operating profile; historical measurements retain their own recorded configuration.

The demonstrated hardware is the 4~GB Arduino UNO Q, combining a Qualcomm QRB2210 Linux microprocessor with an STM32U585 microcontroller \cite{unoq}. CPU-based ONNX inference, ROS~2 Jazzy adapters, image-space advisory, logging and a browser dashboard execute on the Linux side. ONNX Runtime provides the cross-platform inference interface used for the selected exports \cite{onnxruntime}. Arduino App Lab supports the sketch/application workflow; Router Bridge transfers bounded guidance symbols to the MCU. The implementation does not use App Lab AI Bricks and does not demonstrate GPU/NPU inference.

\begin{figure}[htbp]
\centering\includegraphics[width=.86\linewidth]{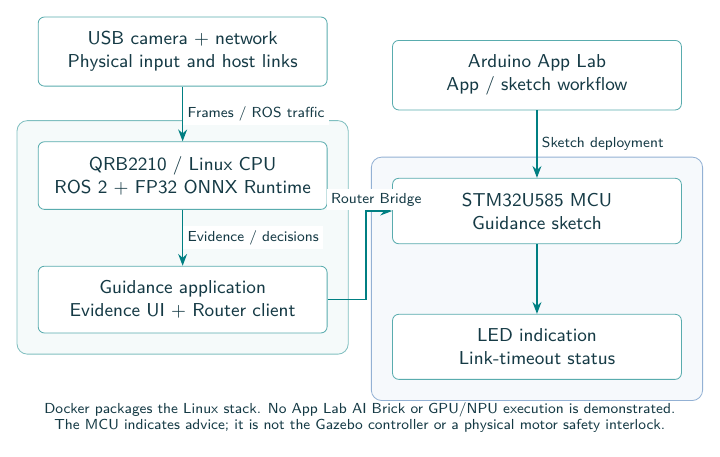}
\caption{UNO Q infrastructure mapping. Linux inference and ROS communication are separate from the sketch's indication role. The App Lab workflow does not imply use of a managed AI Brick.}
\label{fig:unoq}
\end{figure}

An Intel NUC runs Gazebo Harmonic, the vessel model, simulated GPS/IMU, localization and the final ROS-to-Gazebo thrust adapter. Docker provides process/environment packaging on both hosts. The selected graph uses ROS domain~43 and distinct simulation command topics. Physical sensor and motor profiles are not interchangeable with this graph. The camera is a Logitech C525 USB webcam; the current profile requests $320\times240$ acquisition and selects a newest frame once every two seconds.

\begin{figure}[htbp]
\centering\includegraphics[width=\linewidth]{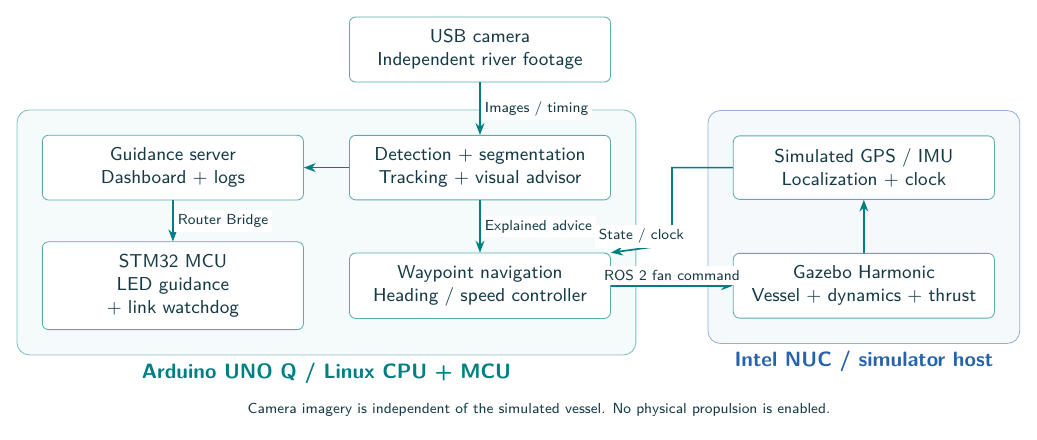}
\caption{Overall compute and robotics partition. The camera's independent footage is not rendered from the simulated pose. The board executes vision and guidance; the external host supplies simulation and sensor feedback. MCU indication is distinct from propulsion control.}
\label{fig:system}
\end{figure}

\subsection{Data contracts and separation of authority}
The camera publishes an image with a ROS header and a matching frame-timing record. The perception worker rejects duplicate/old input and runs the detector followed by the segmenter on the same selected image. Its JSON evidence packet includes image dimensions, source sequence, timestamps, detections, lane summaries, inference timing and freshness. Pixel overlays use separate image topics. The detector's input shape and class order are explicitly specified.

The visual-advisor process associates boxes and publishes tracked evidence and an explained advisory. Neither a neural model nor the visual advisor publishes motor PWM. A separate navigator interprets permitted advisory directions against a route and fresh simulated state. A separate controller produces normalized thrust commands, and a final adapter applies command checks and converts to simulated force. This separation makes errors localizable, but cannot compensate for a confidently incorrect image interpretation.

\begin{table}[htbp]\centering\small
\caption{Principal data interfaces in the selected graph. JSON is carried in ROS String messages.}
\begin{tabularx}{\linewidth}{>{\raggedright\arraybackslash}p{.31\linewidth}YY}\toprule
Interface & Producer and content & Consumer / purpose\\\midrule
Camera image stream & Camera; image + ROS header & Latest-frame inference; valid calibration is not established\\
Frame-timing stream & Camera/replay; source time, epoch & Matched time base for diagnostic motion\\
Perception evidence & Perception; boxes, lane summaries, timing & Tracking and visual policy\\
Tracked observations & Tracker; IDs, acceptance, motion fields & Dashboard and available journal subscribers\\
Explained advisory & Advisor; direction, reasons, components & Navigator, dashboard and guidance bridge\\
Navigation command & Navigator; atomic heading/speed/enable & Heading/speed controller\\
Differential-thrust command & Controller; normalized differential thrust & NUC fan adapter and Gazebo\\
\bottomrule\end{tabularx}\label{tab:interfaces}
\end{table}

\section{Visual perception and evidence fusion}
\subsection{Model contracts}
Both learned exports use static batch-one $[1,3,256,320]$ float32 input. YOLO11n is fine-tuned for person, vessel, pier and buoy. Runtime preprocessing letterboxes while preserving aspect ratio, converts BGR to RGB and scales by $1/255$. Raw output has shape $[1,8,1680]$ for four box coordinates and four class scores at 1,680 candidates. Finite candidates above 0.40 confidence are ranked, at most 256 enter class-agnostic non-maximum suppression at IoU~0.45, and at most 24 boxes are retained. Coordinates are mapped back to source pixels. There is no instance-mask output in the selected four-class detector, even though the general decoder also supports older segmentation exports.

The semantic model uses an ImageNet-initialized MobileNetV3-Small encoder with four scale taps. A $1\times1$ projection reduces the deepest representation to 64 channels. Three top-down blocks resize by nearest neighbor, add a lateral $1\times1$ projection, and refine with $3\times3$ convolution, batch normalization and ReLU. A $3\times3$ three-class head and nearest-neighbor final resize produce water/obstacle/sky logits. This architectural upsampling is distinct from bilinear input resizing. RGB preprocessing uses ImageNet means $(0.485,0.456,0.406)$ and standard deviations $(0.229,0.224,0.225)$.

\begin{table}[htbp]\centering\small
\caption{Selected learned models. ONNX is a serialization format; ONNX Runtime is the board inference engine. Decimal MB are file sizes, not RAM.}
\begin{tabularx}{\linewidth}{p{.22\linewidth}YY}\toprule
Property & Detector & Segmenter\\\midrule
Architecture & YOLO11n bounding-box detector & MobileNetV3-Small + 64-channel light FPN\\
Training framework & PyTorch / Ultralytics & PyTorch / torchvision\\
Class order & person, vessel, pier, buoy & water, obstacle, sky\\
Deployment & FP32 ONNX opset 17 & FP32 ONNX opset 17\\
Input & $[1,3,256,320]$ & $[1,3,256,320]$\\
Output & $[1,8,1680]$ raw predictions & $[1,3,256,320]$ logits\\
Parameters & 2,582,932 (fused detector) & 1,082,467 (trainable)\\
Artifact size & 10.472507 MB & 4.344661 MB\\
CPU runtime & ONNX Runtime / CPU provider & ONNX Runtime / CPU provider\\
Benchmark threads & 2 intra-op; 1 inter-op & 2 intra-op; 1 inter-op\\
\bottomrule\end{tabularx}\label{tab:models}
\end{table}
The host evaluation environment records ONNX Runtime~1.30.0. An earlier UNO Q image records version~1.20.1, but the five-minute benchmark metadata does not independently identify its runtime version. The two-thread setting in Table~\ref{tab:models} describes that benchmark, not all replay or board profiles.

\begin{figure}[htbp]\centering
\includegraphics[width=.88\linewidth]{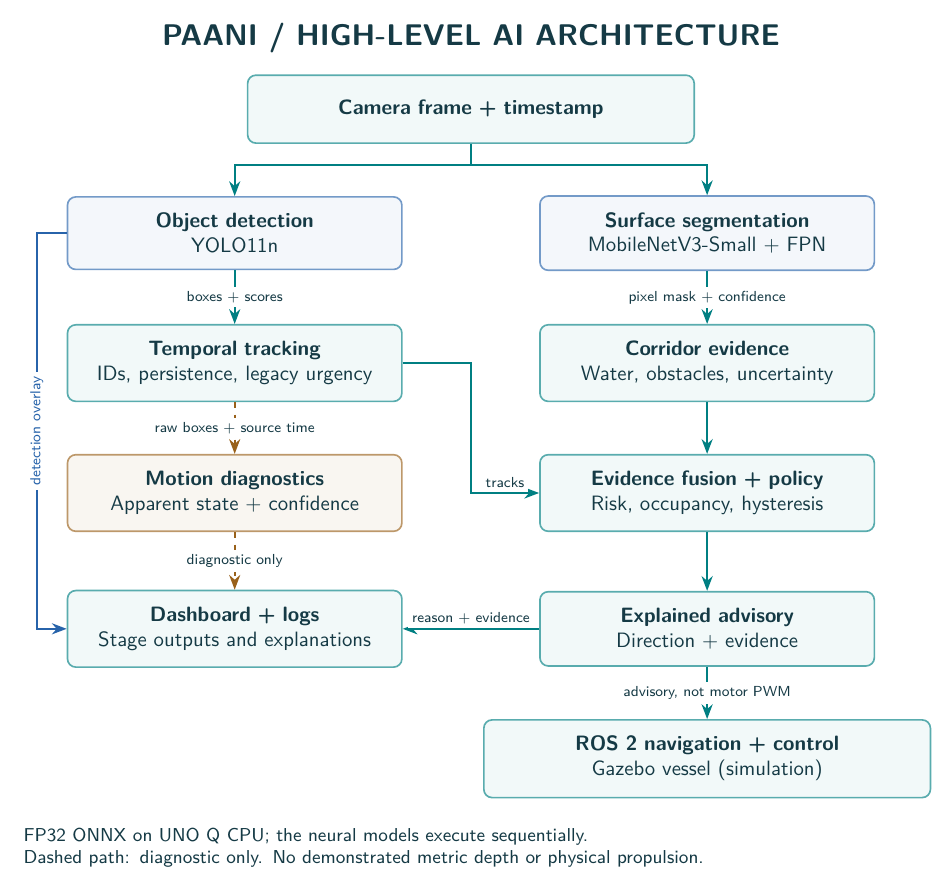}
\caption{High-level AI architecture. The two learned branches express dependencies, not concurrent execution. The new apparent-motion states are a diagnostic side branch. Legacy urgency remains an input to the existing risk model. Detailed contracts appear in Tables~\ref{tab:interfaces} and~\ref{tab:models}.}
\label{fig:ai}
\end{figure}

\subsection{Pixel-to-corridor reduction}
Let $p_c(u,v)$ be the three-class softmax probabilities. The predicted label is $\hat y=\arg\max_c p_c$ and pixel confidence is $\max_c p_c$. The evaluated region begins at normalized height 0.30. Overlapping horizontal intervals are left $[0.03,0.39]$, center $[0.31,0.69]$, and right $[0.61,0.97]$. For corridor $j$, non-water evidence $N_j$ is the fraction of pixels not labeled water and $U_j=1-\operatorname{mean}(\max_c p_c)$ is the uncertainty proxy. Clear width is the longest contiguous run of columns containing at least 60\% water, normalized by corridor width; it is logged but is not an independent metric clearance certificate.

An HSV/corridor fallback can produce labeled visual output when the learned segmenter is unavailable. It declares itself non-learned and navigation-ineligible. The advisor then substitutes default water components rather than using its mask as validated free-space evidence. This fallback exclusion does not make the overall policy fail-safe: low obstacle evidence can still allow STRAIGHT.

\subsection{Temporal association and legacy urgency}
The tracker operates on normalized boxes with bounded work. Greedy IoU association uses threshold 0.25, a 3.0-s hold in the selected profile, and two-hit confirmation. A box in the immediate-danger region can be accepted without waiting for confirmation. Display-box smoothing retains 0.35 of the previous box; confidence smoothing retains 0.65 of the previous score. Labels are preserved for audit and mapped to a compact hazard vocabulary. The currently trained ontology has no dedicated litter, log or vegetation class.

The existing urgency path uses apparent growth of square-root normalized box area and movement toward the image center. Each positive rate is scaled and saturated, with full-scale rates 0.16 and 0.20~s$^{-1}$ respectively. Urgency is the maximum of these scores. Held observations decay and missed detections do not constitute new motion samples. Confirmation, immediate-danger acceptance and prediction status remain explicit fields rather than hidden changes to the detector score.

\subsection{Diagnostic apparent-motion estimator}
For source-timed observations $(t_i,b_i)$, let $a_i$ be normalized box area and $y_i=\log\sqrt{a_i}$. The least-squares apparent-scale rate is
\begin{equation}
\hat\beta=\frac{\sum_i(t_i-\bar t)(y_i-\bar y)}{S_{tt}},\qquad S_{tt}=\sum_i(t_i-\bar t)^2.
\end{equation}
Its units are inverse seconds, not metric speed. A heuristic margin is
\begin{equation}
m=\max\left(0.01,\;3\sqrt{\frac{\sum_i[y_i-\bar y-\hat\beta(t_i-\bar t)]^2}{(n-2)S_{tt}}}\right).
\end{equation}
Approaching requires $\hat\beta-m>0.04$; receding requires $\hat\beta+m<-0.04$; approximately steady requires the complete interval to lie in $[-0.04,0.04]$. Other cases are unknown. The interval is not a statistically calibrated confidence interval.

The estimator stores at most 24 samples over 1.5 source seconds and requires at least five samples over 0.4~s. A confidence-quality product combines duration, sample count, minimum detector confidence and residual margin. Quality below 0.25 yields unknown; reported non-unknown confidence is capped at 0.9. Frame-edge clipping, weak detections, ambiguous association, abrupt box-shape changes, invalid/reversed time, replay epoch changes and gaps above 0.75~s reset evidence.

ROS stamps support freshness, while matched media timestamps or explicitly labeled frame-index/FPS fallback support replay motion. A webcam observing a monitor uses its capture clock and sees projected monitor changes; it cannot recover the original river range. Camera translation, zoom, object pose and identity errors can imitate scale change. No image stabilization or camera-motion compensation is implemented. The four new state labels are diagnostic only and cannot lower risk or authorize navigation.

\subsection{Late fusion, policy and explanation}
For each accepted box, the advisor computes a bounded severity from confidence, normalized bottom-edge proximity and vertical extent. Corridor overlap weights its obstacle contribution. Multiple contributions combine as $\E(a,b)=1-(1-a)(1-b)$, producing bounded obstacle evidence $O_j$ and legacy urgency evidence $M_j$. The lane score is
\begin{equation}\label{eq:risk}
R_j=\min(1,0.55N_j+0.65O_j+0.20M_j+0.15U_j).
\end{equation}
The weights are engineering defaults, not learned probabilities or empirically optimized collision costs. Center/side blocking thresholds are 0.48/0.58. Persistent lower-image occupancy, a center-vessel area threshold of 0.04, permitted-side checks, and center-edge versus center-core incursion can override score-only behavior. A turn may be retained until two distinct clear frames occur, preventing timer republishes from being mistaken for fresh visual evidence.

The final advisory contains direction, reason code, per-lane components, thresholds, occupancy, contributing track IDs and age. Explanations are regenerated after hysteresis and error overrides. ``Contributing'' means overlap with an accepted lane ROI, not a counterfactual proof that an individual box caused the action. Stale input yields STOP; malformed or missing evidence yields STOP or NO\_DATA according to the validation branch. None of these checks identifies all visually unusable but syntactically valid images.

\section{Robotics implementation}
\subsection{Simulation and state estimation}
The evaluated simulation environment is an open approximately $208.699\times126.000$~m lake, with a 59-waypoint repeating route of approximately 465.443~m. It contains no internal bank geometry corresponding to the independent footage. The vehicle model is a provisional twin-fan catamaran. Its surface-dynamics plugin applies body-axis linear/quadratic drag
\begin{equation}F_d(v)=-c_1v-c_2|v|v,
\end{equation}
vertical buoyancy/restoring force
\begin{equation}F_z=\clip\{mg+k_z(z_e-z)-d_z\dot z,0,F_{\max}\},
\end{equation}
and roll/pitch restoring and damping torques plus yaw drag. The lumped mass is 12.4~kg; the equilibrium height is 0.177~m. Coefficients are provisional, not fitted hydrodynamic parameters. The simulator is not CFD and has no validated current, wave or wind model.

A sensor adapter supplies ROS GPS and IMU messages. A 20-Hz local IMU-oriented EKF, GPS-to-local-frame transformation, and a 20-Hz global GPS-position/IMU-heading EKF provide local ENU state. These estimates are not Gazebo ground truth. GPS-derived horizontal velocity can remain nonzero while stationary; the controller includes a stopped-speed guard rather than treating that estimate as evidence of actual motion. The physical MPU6050 bridge marks orientation unavailable, so replacing simulated sensors with that bridge does not automatically supply absolute heading.

\subsection{Route-conditioned guidance and control}
The selected navigator is a camera-simulation route follower, not the historical M10 YAML mission manager. It retains the current waypoint during visual turns, advances within a 1-m acceptance radius, and repeats the route. STRAIGHT uses the waypoint bearing $\psi_w=\operatorname{atan2}(y_w-y,x_w-x)$. A LEFT/RIGHT advisory instead selects current yaw plus/minus $25^\circ$; slight turns use $10^\circ$. Turn speed is half the configured 0.15~m/s cruise. Other advisory states disable motion. Thus this interface is a heading heuristic, not a planned metric detour around the observed object.

Navigation requires the simulation enable condition, domain~43, advancing simulation time and fresh pose/sensors/advice. Visual age allowance is 3.5~s in this low-rate profile; the navigation freshness check also bounds time since advice receipt. The command is atomic: heading, speed, enable, frame and validity are checked together. Clock pause/reset, stale/future/replayed commands and nonfinite values are rejected by the applicable layers.

Heading/speed control uses proportional speed feedback with feed-forward and PD heading correction, then differential thrust mixing, saturation, reverse suppression and slew limiting. In the selected profile it runs at 10~Hz with maximum target speed 0.3~m/s, normalized forward limit 0.08, steering limit 0.04, and slew rate 0.15~s$^{-1}$. Gains are heading $K_p=0.08$, $K_d=0.05$, speed $K_p=0.08$ and speed feed-forward 0.16. The NUC adapter converts normalized commands using a configured software cap of 75~N per fan and zeros output after a 0.5-s wall-clock timeout. This is not a measured physical thrust capability. These commands act on Gazebo; no physical ESC interface is demonstrated.

\subsection{MCU, observability and experimental extensions}
The guidance server exposes camera/model evidence and decision explanations to a browser. Perception has its own bounded observation journal. Tracked detections and decisions remain ROS/dashboard telemetry; optional journal subscribers can record them with bounded queues and rotation, whose drop telemetry must be considered when interpreting completeness. The selected camera-simulation launch disables the optional survey and benchmark nodes, including the benchmark decision journal. Collector CSVs and per-process logs should therefore not be conflated with an always-enabled event journal. MCU LED guidance has a communication timeout after an established stream; it is an operator indicator, not an independent motor interlock.

A separate metric prototype is outside the evaluated configuration. With valid camera intrinsics and transforms, a ray $r=RK^{-1}[u,v,1]^\top$ can intersect a measured water plane at $p=o+r(z_w-o_z)/r_z$, subject to horizon/range rejection. Its local grid is $100\times100$ cells at 0.2~m resolution; free and occupied evidence expire after 0.8 and 2.0~s. Candidate motion is checked with swept hull footprints over a 3-s horizon at 0.2-s increments. This prototype requires calibrated geometry and navigation-kernel dependencies; it is not the active image-space advisor and is not demonstrated by the present metrics. Sampling payload actuation and physical propulsion likewise remain outside the evaluated configuration.

\section{Datasets and training}
\subsection{Data selection, ontology and split integrity}
The detector experiment prepares 26,236 WaterScenes RGB images: 10,000 selected from the official training pool, all 10,824 official validation images, and all 5,412 official test images. Selection uses seed~42 and prioritizes images containing buoy, sailor or kayak before filling the remainder. All mapped boxes in selected images are retained. Source sailor maps to person; ship, boat, vessel and kayak map to vessel; pier and buoy retain separate classes. The evaluated ontology contains four navigation-relevant classes. It is intended for visible navigation-relevant categories, not pollution identification or comprehensive person-in-water detection.

\begin{table}[htbp]\centering\small
\caption{Prepared WaterScenes sample and annotation counts. The selected subset is not the entire public dataset.}
\begin{tabular}{lrrrrr}\toprule
Split & Images & Person & Vessel & Pier & Buoy\\\midrule
Training &10,000&5,594&9,484&28,301&11,538\\
Validation &10,824&1,609&11,254&24,334&3,320\\
Test &5,412&835&5,679&12,186&1,680\\\bottomrule
\end{tabular}\label{tab:data}
\end{table}

Preparation validates normalized boxes and rejects overlapping image IDs between splits. These checks do not establish location independence or absence of near-duplicate scenes. MaSTr1325 supplies 1,127 training and 198 validation images from a seeded image-level 0.15 split. Original obstacle label~0 maps to PAANI~1, water~1 to PAANI~0, sky remains~2, and unknown boundaries map to ignore label~255. No independent segmentation test set or sequence-held-out guarantee exists. The segmentation result therefore measures this split, not general river performance. Other datasets listed in early project plans, including LaRS, were not used in the selected completed runs.

\subsection{Optimization and augmentation}
Training used an NVIDIA RTX A2000 12~GB, Python~3.12, PyTorch~2.5.1 with CUDA~12.1, torchvision~0.20.1, Ultralytics~8.4.150 and Albumentations~2.0.8. Both runs used CUDA automatic mixed precision; this is compatible with subsequently selecting FP32 deployment artifacts.

YOLO11n began from pretrained weights and completed 60 epochs at square 320-pixel input, batch~32, using AdamW with initial learning rate 0.001 and weight decay 0.0005. The cosine schedule had final learning-rate fraction 0.01 and three warmup epochs. Early stopping patience was~12 but the full 60 epochs completed. Loss coefficients were box~7.5, class~0.5 and distribution focal~1.5. Augmentations included HSV perturbation, translation, scaling, horizontal flip and mosaic; mosaic was disabled for the last ten epochs. The final deployment is rectangular $320\times256$, so checkpoint and exported-model results are reported separately.

The segmenter completed 50 epochs, batch~16, $320\times256$ input, AdamW learning rate $3\times10^{-4}$, weight decay $10^{-4}$ and a 50-epoch cosine schedule. The objective was cross-entropy plus Dice with equal coefficient, ignoring label~255 in the loss/IoU path. Training used horizontal flip, color jitter and synthetic glare. The decoder width was~64. The best checkpoint, selected by validation mIoU, occurred at one-based epoch~31; the final epoch is not substituted for the selected best model. Detailed hyperparameters appear in Appendix~\ref{app:hyper}.

\subsection{Training records and determinism limits}
The saved detector CSV contains all 60 epochs, including training and validation box/class/DFL loss, precision, recall and mAP. The segmenter JSONL contains all 50 epochs, training loss, class IoU, mIoU and edge F1. A segmentation validation-loss curve was not recorded. Segmentation checkpoints contain model, optimizer and scheduler state plus arguments, but not every RNG/AMP state. The distributed detector best/last checkpoints have stripped optimizer state and no scheduler entry, so they are not full training-resume snapshots. The glare augmentation creates an unseeded NumPy generator per call; a global seed therefore does not guarantee a bit-identical rerun. No multi-seed confidence intervals or systematic hyperparameter search are available.

\begin{figure}[htbp]\centering\includegraphics[width=\linewidth]{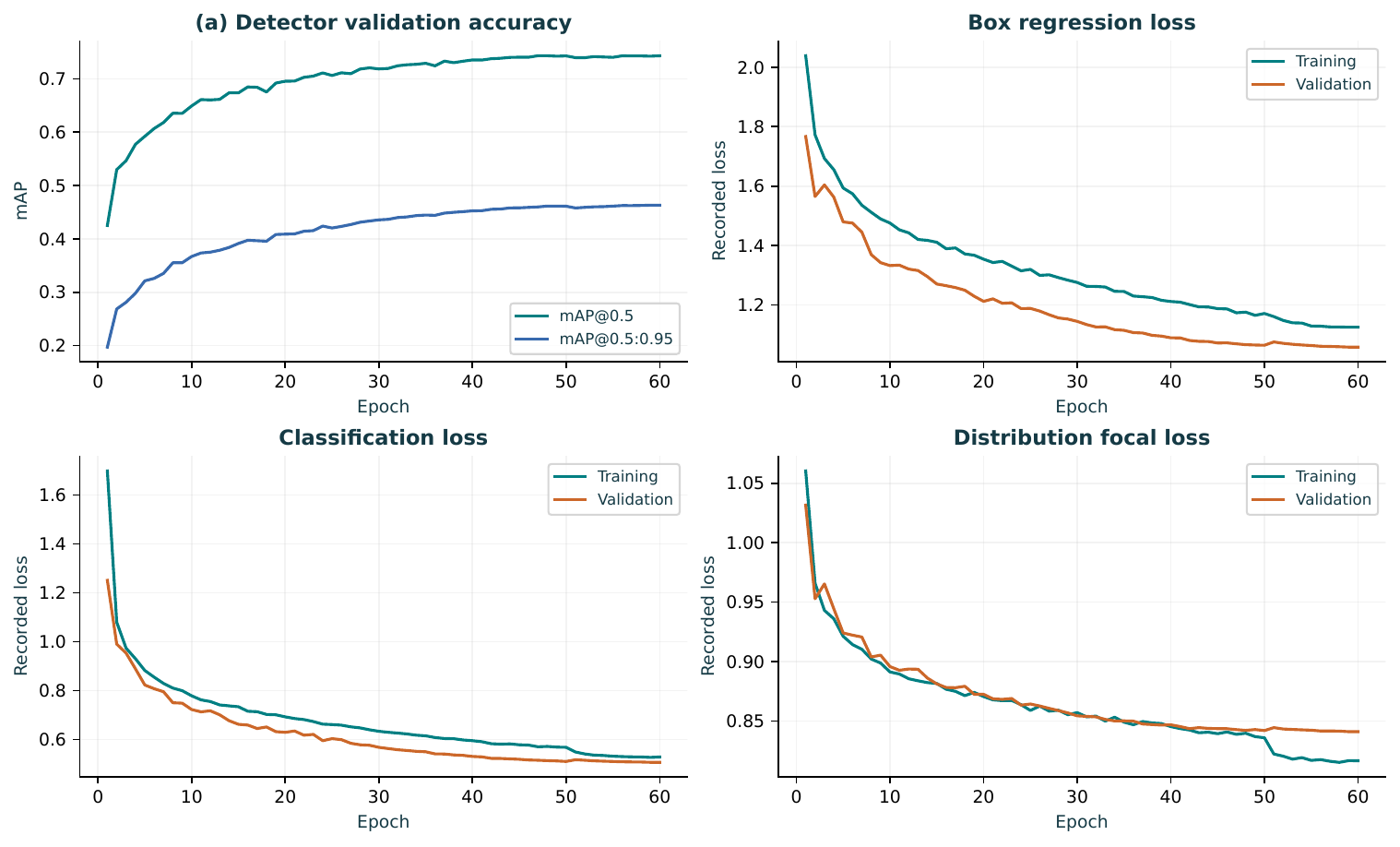}
\caption{Complete detector training history from 60 recorded epochs. Accuracy and loss are dataset-fitting diagnostics, not closed-loop navigation metrics. The evaluation geometry of the selected rectangular export is reported separately.}
\label{fig:detcurves}\end{figure}
\begin{figure}[htbp]\centering\includegraphics[width=\linewidth]{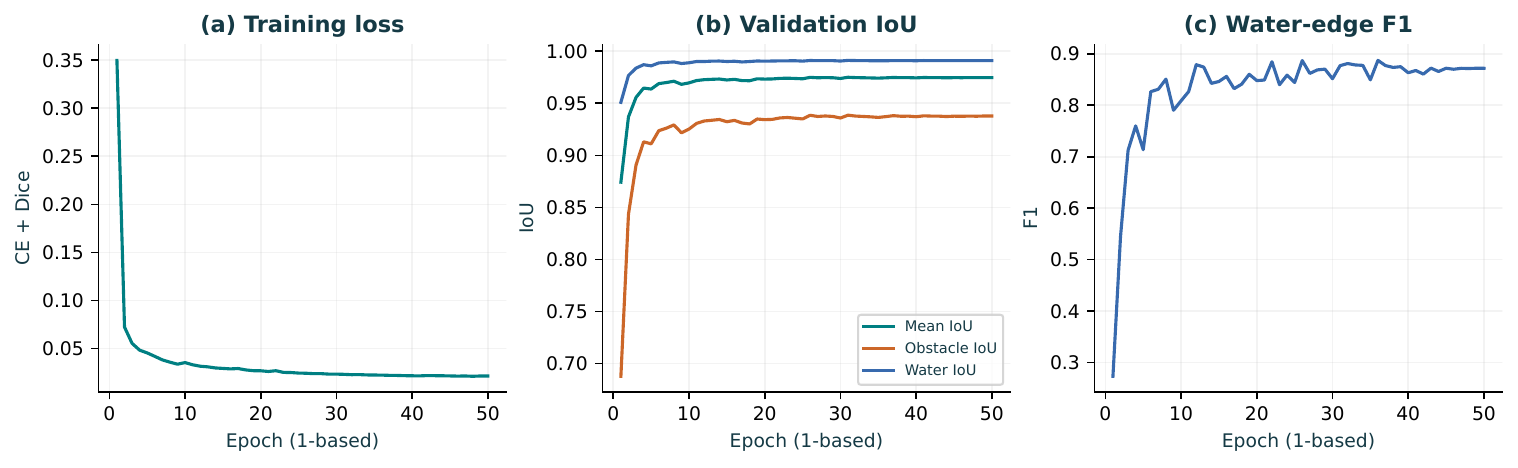}
\caption{Complete 50-epoch segmentation history, with epoch labels converted from zero-based logs to one-based display. No validation-loss series is inferred. High water/sky IoU should be considered together with obstacle and boundary quality.}
\label{fig:segcurves}\end{figure}

\subsection{Export and precision selection}
Static ONNX opset~17 exports use float32 input/output. Structural and numerical checks reported maximum absolute differences of approximately $7.63\times10^{-4}$ for the detector and $4.12\times10^{-4}$ for the segmenter, below the recorded $10^{-3}$ tolerance. Such agreement checks test export consistency, not semantic accuracy. The FP32 artifacts were evaluated separately on labeled data.

The reported INT8 candidate uses training-only calibration. The procedure used 200 sorted training images, MinMax calibration, QDQ nodes and signed INT8 tensors. The file shrank from approximately 4.34 to 1.44~MB, but mIoU fell from 0.9750 to 0.7303 and edge F1 from 0.8769 to 0.2318. A host CPU microbenchmark also increased from approximately 6.969 to 10.651~ms. Those host timings are not UNO Q power or latency measurements. A separate candidate calibrated on 198 validation images is not the candidate reported in this table. The tested training-calibrated INT8 configuration was rejected; no quantization-aware-training result or complete detector/segmenter precision matrix is claimed.

\section{Evaluation protocol}
We separate four evidence classes. First, labeled model evaluation measures prediction quality on the recorded splits. Second, historical offline scenarios exercise model-to-advisory interfaces using four still images and a generated black image. Each still is repeated eight times at controlled 10-Hz source timestamps; these are not independent video observations. Third, generated box trajectories and regression tests examine estimator state logic. Fourth, recorded physical-board integration logs measure timing and software health. Results from these classes cannot be substituted for one another.

Detector checkpoint mAP uses the official untouched test list; the rectangular ONNX validation report uses a separate 101-point evaluator on the official validation list. mAP@0.5:0.95 averages across IoU thresholds. Segmentation mIoU averages water/obstacle/sky IoU over the 198-image validation split. The project's water-edge F1 uses a four-neighbor boundary extraction and two-pixel matching tolerance, not the MODS obstacle protocol; its ignore treatment differs from the IoU path. Scores should therefore not be compared directly with literature leaderboards.

The principal board run comprises 300 one-second polls over five minutes and 151 distinct inference results. Inference percentiles are computed from distinct results; other telemetry uses available polls. Percentiles use nearest rank, while an even-sample median averages the middle pair. CPU is whole-board utilization; cgroup memory includes charged cache/kernel memory and is not model-only RSS. No energy meter, battery-endurance experiment or labeled motion dataset was used. Historical tests have different scopes and operating configurations; their counts are not pooled as end-to-end system acceptance.

\section{Results}
\subsection{Labeled perception performance}
\begin{table}[htbp]\centering\small
\caption{Detection results. Separate artifacts and evaluation splits are not interchangeable.}
\begin{tabularx}{\linewidth}{Ylrr}\toprule
Artifact & Split & mAP@0.5 & mAP@0.5:0.95\\\midrule
Selected PyTorch checkpoint & Validation &0.743199&0.463867\\
Selected PyTorch checkpoint & Test &0.738781&0.457888\\
Rectangular FP32 ONNX & Validation &0.736668&0.456929\\\bottomrule
\end{tabularx}\label{tab:detresults}
\end{table}
\begin{table}[htbp]\centering\small
\caption{Per-class detector AP@0.5:0.95. The results reveal category variation hidden by a single aggregate.}
\begin{tabular}{lrrr}\toprule
Class & Checkpoint val. & Checkpoint test & ONNX val.\\\midrule
Person &.366075&.365398&.359960\\
Vessel &.609288&.609214&.599580\\
Pier &.421294&.412297&.418510\\
Buoy &.458810&.444641&.449664\\\bottomrule
\end{tabular}
\end{table}
\begin{table}[htbp]\centering\small
\caption{Segmentation validation on 198 images. INT8 is the rejected PTQ candidate.}
\begin{tabular}{lrrrrr}\toprule
Artifact & mIoU & Water IoU & Obstacle IoU & Sky IoU & Edge F1\\\midrule
PyTorch &.975038&.991244&.938622&.995249&.876991\\
FP32 ONNX &.975037&.991245&.938620&.995246&.876919\\
INT8 ONNX &.730281&\multicolumn{3}{c}{Not used for deployment}&.231788\\\bottomrule
\end{tabular}\label{tab:segresults}
\end{table}
The checkpoint test detector score is lower for person, pier and buoy than vessel (Table~\ref{tab:detresults} and the class breakdown). The semantic model's aggregate is high, but the obstacle and boundary measures and the image-level split limitation remain important. The available evidence does not isolate the gain due to each augmentation, architecture choice or data-selection rule.

\subsection{Stage outputs and qualitative behavior}
\begin{figure}[htbp]\centering\includegraphics[width=\linewidth]{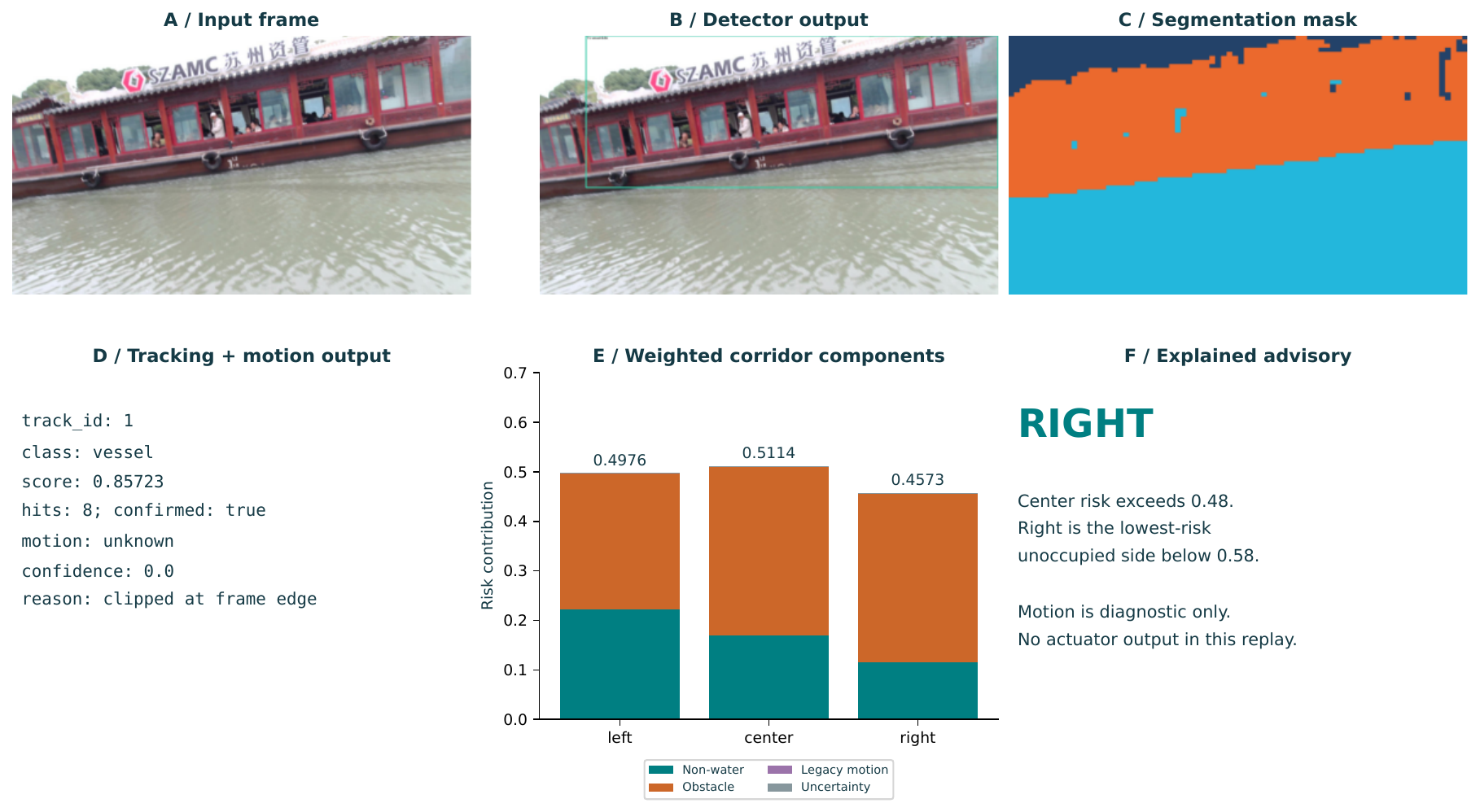}
\caption{Per-stage sample outputs for WaterScenes image 01703. A--C are saved image/model outputs; D--F render matching recorded telemetry. The mask is predicted, not ground truth. The Windows-host replay used threshold 0.25, eight repeated still observations and controlled timestamps; the board uses threshold 0.40. Neither physical motion nor field generalization is established.}
\label{fig:stages}\end{figure}

Figure~\ref{fig:stages} follows one saved scene through input, detection, segmentation, tracking, fusion and advice. A vessel confidence of 0.85723 coexists with unknown motion because its raw box extends beyond the frame edge. This is an example of keeping detector confidence separate from validity of temporal evidence. Center risk is 0.5114, while the unoccupied right corridor has risk 0.4573; the policy emits RIGHT. These values come from the same final replay observation, not independently chosen outputs.

\begin{table}[htbp]\centering\small
\caption{Final historical offline decisions. L/C/R are corridor scores, not probabilities. All five stale probes produced STOP after the recorded 9.3-s age.}
\begin{tabularx}{\linewidth}{lrrrlY}\toprule
Input & L & C & R & Advice & Interpretation\\\midrule
47266 &.3413&.2941&.1611&STRAIGHT&Center below blocking threshold\\
01703 &.4976&.5114&.4573&RIGHT&Center blocked; lower-risk permitted right side\\
27011 &.3367&.4716&.6280&STRAIGHT&Center remains below 0.48\\
47357 &.6615&.6025&.4489&STOP&Occupancy prevents a permitted safer turn\\
Black &.0288&.0004&.0029&STRAIGHT&Unusable-input failure, not safe water\\\bottomrule
\end{tabularx}\label{tab:cases}
\end{table}

\begin{figure}[htbp]\centering\includegraphics[width=\linewidth]{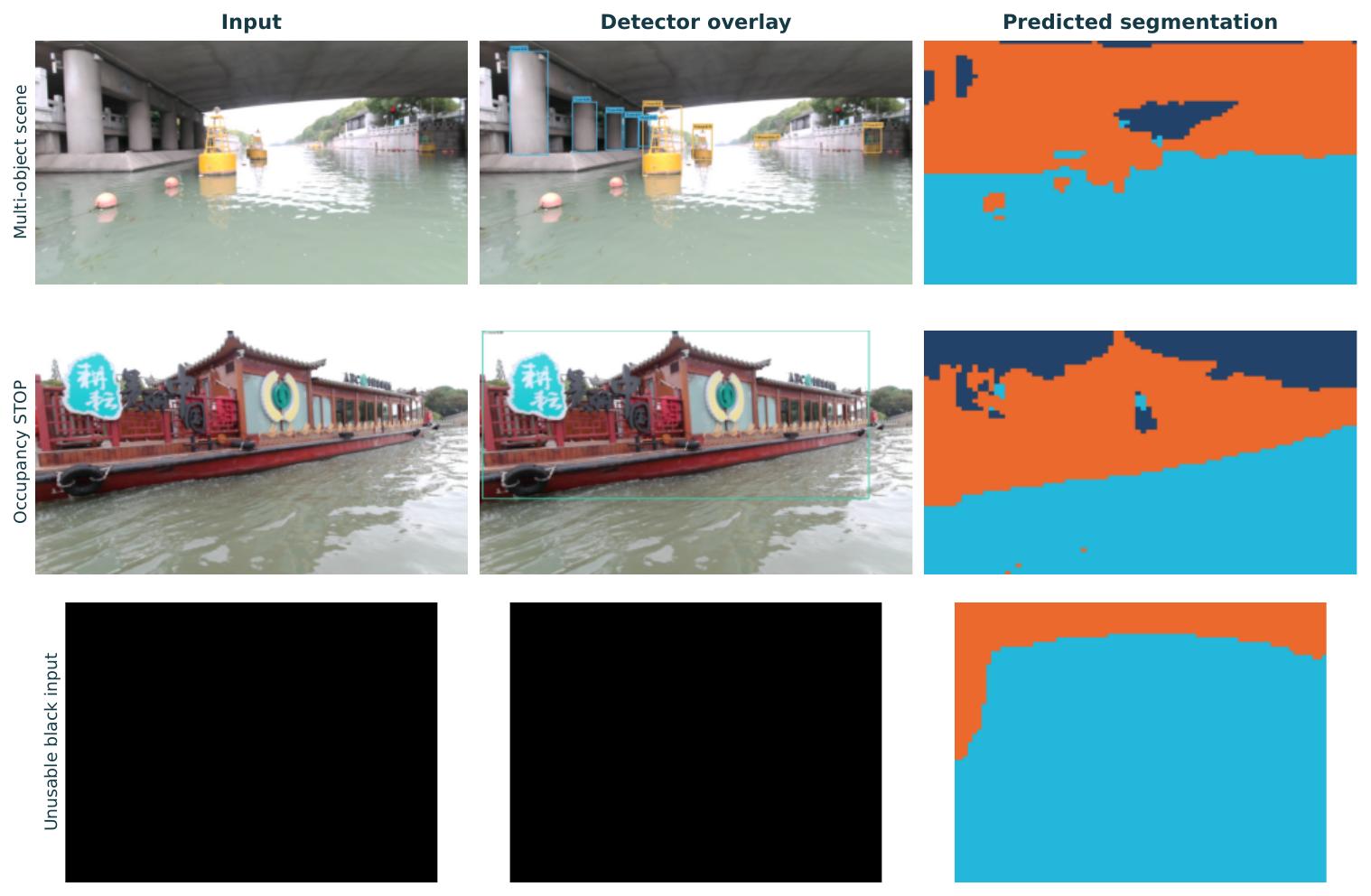}
\caption{Saved multi-object, occupied-corridor and black-input examples. Cyan denotes predicted water, orange obstacle/environment, and dark blue sky. These selected illustrations are not a held-out qualitative test set. The last row exposes an invalid-input failure rather than a successful negative control.}
\label{fig:qualitative}\end{figure}

The black image was labeled approximately 97.56\% water in the evaluated ROI and produced STRAIGHT. A syntactically valid, recent image can therefore pass freshness and tensor checks while containing no useful scene evidence. This is consistent with the broader distinction between classifier confidence, calibration and explicit out-of-distribution detection \cite{guo2017,hendrycks2017}. The explanation faithfully reports the policy's inputs, but faithfulness is not correctness. This result rules out describing the current system as safe under arbitrary camera failure.

\subsection{Temporal diagnostics and cadence mismatch}
Generated expansion/contraction/steady trajectories test the implemented state logic, while clipped boxes, timing gaps and ambiguity test rejection. Such tests do not label physical approach/recession in river video. At 0.5~Hz, adjacent selected frames are approximately 2~s apart, exceeding the 0.75-s permitted source gap and the 1.5-s history window. Thus at most one usable observation is retained per new update, below the minimum five. Implementing the four-state estimator is not the same as demonstrating it at the deployed cadence. The recorded motion-deployment session reported unknown labels throughout.

The legacy tracker can still associate across the configured 3-s hold and compute its older urgency scores. Those signals and the new diagnostic state must not be merged in interpretation. Increasing rate alone is also insufficient evidence of physical validity: camera motion, clipping and unmodeled changes remain confounders.

\subsection{UNO Q execution and robotics outputs}
\begin{table}[htbp]\centering\small
\caption{Recorded five-minute two-thread UNO Q benchmark. Timing and thermal observations do not constitute power or long-duration qualification.}
\begin{tabularx}{\linewidth}{Yr}\toprule
Metric & Value\\\midrule
Polling samples / distinct inference results &300 / 151\\
Pipeline latency median / p95 / maximum (ms)&467.842 / 580.341 / 643.857\\
Detector median / segmenter execution median (ms)&254.408 / 165.593\\
Effective inference cadence (Hz)&0.500\\
Measured capture rate, approximately (FPS)&2.58\\
Frame age at completion median / p95 (s)&0.6838 / 0.8888\\
Whole-board CPU median (\%)&46.32\\
Container memory median (MiB)&1665.7\\
cpuss0 temperature median / maximum ($^\circ$C)&45.3 / 50.3\\
Measured power / battery endurance&Unavailable\\\bottomrule
\end{tabularx}\label{tab:board}
\end{table}

\begin{figure}[htbp]\centering\includegraphics[width=\linewidth]{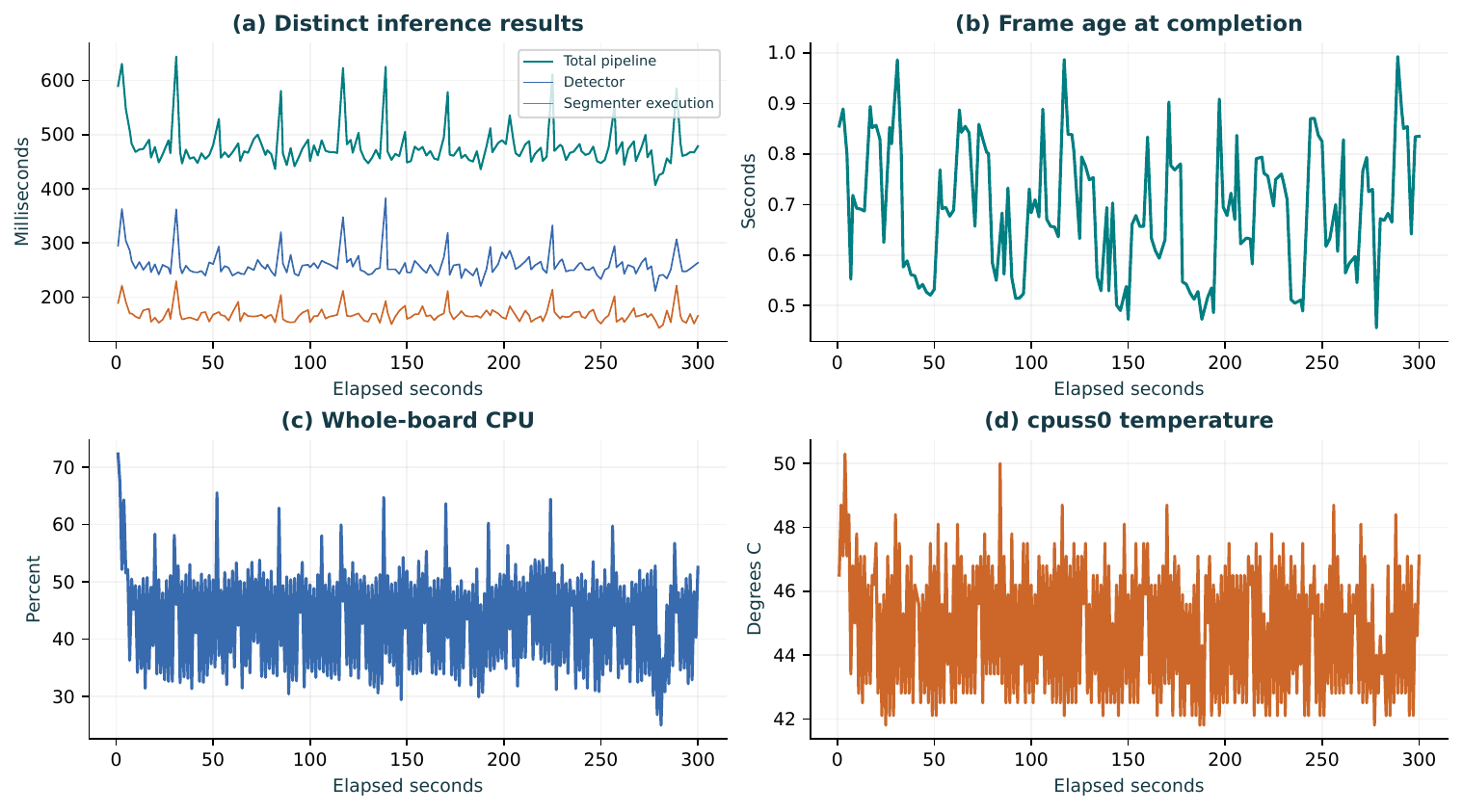}
\caption{Replotted raw operational telemetry. Inference/age use 151 distinct results; CPU and temperature use 300 polls. Detector timing includes its runtime preprocessing/decoding, while segmenter timing is the recorded session execution interval; their medians need not sum to total pipeline median.}
\label{fig:board}\end{figure}

All 300 sampled API requests succeeded and reported learned segmentation. No sampled stale advisory was present; this does not exclude transients between polls. CPU quota-throttling and OOM-kill counters did not increase; quota counters are not a thermal-throttling measurement. Compared with the earlier trained one-thread operational run, the two-thread run's median total latency decreased from 539.6 to 467.8~ms. Scenes, warm-up, background work and runtime configurations were not fully controlled, so this is not a causal ablation. The model pair's 14.817168-MB disk footprint is much smaller than the complete runtime memory charge.

A recorded integration snapshot at 15:45:19.863 UTC on September~13 shows SLIGHT\_LEFT, navigation enabled at waypoint index~12, and applied normalized left/right fan commands 0.00000/0.02611. This is a Gazebo-interface sample from a different session than Figure~\ref{fig:stages}. Earlier camera/simulation evidence recorded vessel displacement over about 27~s with zero physical motor publishers. Neither record demonstrates a complete camera-guided lap or avoidance of a visible real obstacle. Historical M10 route acceptance is a separate baseline and is not pooled with these results.

\section{Discussion and limitations}
\subsection{What is supported}
The evidence supports task-trained model execution on a constrained CPU, traceable conversion from pixel/box evidence to an advisory, and integration with a modular simulator/control stack. The design permits a developer to distinguish a model error, temporal rejection, occupancy override or stale-message condition. Separating model contracts, risk components and control authority is useful even when a failure is discovered: it identifies a testable intervention rather than hiding the problem behind a confident instruction.

\subsection{What is not supported}
No physical river deployment, water-quality measurement, waste collection, rescue detection, battery-endurance result or calibrated monocular distance is established. The independent video/simulation geometry prevents a closed-loop visual navigation claim. High segmentation mIoU cannot establish small-obstacle recall or collision avoidance. The model confidence and motion-quality fields are not calibrated probabilities, and the black-frame result demonstrates that a strong-looking confidence can be misleading.

The evidence is also limited by selected demonstration scenes, image-level segmentation splitting, lack of multi-seed training, absent cross-river tests and absence of matched ablations. A rigorous next evaluation would hold out complete sequences/sites, include low light, reflections, rain, occlusion and camera disconnection, annotate physical motion with an appropriate reference, and measure coupled-scene task success and failure recovery. These are proposed experiments, not completed results.

\subsection{Reusability, sustainability and deployment implications}
PAANI's reusable element is the interface pattern: timestamped model evidence, bounded state, explicit policy and inspectable outputs. Canal/reservoir inspection or aquaculture observation could retain the software boundaries while changing data and task rules. Ground robotics requires a new traversability representation, not unchanged water labels. Adding sensors or actuators requires calibration, driver integration, operating-envelope tests and hardware safety controls.

On-device inference removes a mandatory cloud-video dependency from the decision path. It does not prove lower net energy consumption; the full demonstration includes an external simulator and power conversion. Sustainability benefits such as broader measurement coverage, lower energy or reduced waste require an explicit operational baseline and measured outcomes. Likewise, ROS modularity and container packaging facilitate adaptation but do not prove fleet-scale reliability or hard real-time behavior.

Before physical actuation, unusable-image rejection and a conservative navigation response must address the black-frame failure, and motion features must be validated at the actual sampling rate. Independent hardware stop paths, adequate range/velocity sensing and field-calibrated dynamics are additional requirements. Explainability is a debugging and accountability mechanism, not a substitute for those controls.

\section{Conclusion}
PAANI demonstrates an on-device, project-trained perception stack with transparent temporal and corridor-based guidance integrated into a ROS~2/Gazebo robotics testbed. The selected FP32 models provide compact deployment artifacts and recorded CPU execution below one second per sampled frame, while the explicit evidence chain makes both decisions and defects inspectable. Its strongest present contribution is a local AI foundation for robotics development, not proven on-water autonomy. The results also illustrate why model quality, timing, diagnostic validity and system safety must be evaluated separately: a high validation score can coexist with a black-image failure, and implemented motion logic can remain unusable at the deployed cadence.

\section*{Code availability}
Code is available at \url{https://github.com/Savio-Cardoz/paani}.

\FloatBarrier
\appendix
\section{Detailed hyperparameters and bounded runtime constants}\label{app:hyper}
\begin{table}[H]\centering\small
\caption{Recorded detector optimization and augmentation configuration. Numeric settings are not a claim of a tuned optimum.}
\begin{tabularx}{\linewidth}{p{.36\linewidth}Y}\toprule
Setting & Value\\\midrule
Epochs / batch / workers &60 / 32 / 4\\
Initialization / input &Pretrained YOLO11n; square 320 training input\\
Optimizer / initial LR / weight decay &AdamW / 0.001 / 0.0005\\
Cosine final LR fraction / warmup &0.01 / 3 epochs\\
Momentum / warmup momentum &0.937 / 0.8\\
Warmup bias LR / nominal batch &0.1 / 64\\
Loss weights: box / class / DFL &7.5 / 0.5 / 1.5\\
Seed / deterministic / AMP &42 / requested / enabled\\
Patience / completed epochs &12 / 60\\
HSV hue / saturation / value &0.015 / 0.5 / 0.3\\
Translation / scale / horizontal flip &0.1 / 0.3 / 0.5\\
Mosaic / close mosaic &1.0 / final 10 epochs\\
Vertical flip / rotation / shear / perspective &0 / 0 / 0 / 0\\
Mixup / cutmix / copy-paste &0 / 0 / 0\\
Recorded auto-augment / erasing fields &randaugment / 0.4 (task-dependent configuration fields)\\
Rectangular training flag &false\\\bottomrule
\end{tabularx}
\end{table}

\begin{table}[H]\centering\small
\caption{Recorded segmentation configuration and selected runtime thresholds.}
\begin{tabularx}{\linewidth}{p{.38\linewidth}Y}\toprule
Setting & Value\\\midrule
Segmenter epochs / batch / workers &50 / 16 / 4\\
Decoder / input &64 channels; $320\times256$\\
Optimizer / LR / weight decay &AdamW / 0.0003 / 0.0001\\
Schedule / loss &Cosine $T_{\max}=50$; CE + Dice (coefficient 1)\\
Ignore label / seed / AMP &255 / 42 / enabled\\
Training loader &Shuffled; incomplete final batch dropped\\
Horizontal flip / color jitter probability &0.5 / 0.5\\
Jitter brightness / contrast / saturation / hue &0.2 / 0.2 / 0.2 / 0.05\\
Synthetic glare / blob / glitter probabilities &0.35 / 0.7 / 0.6\\
Best checkpoint / final epoch &31 (one-based) / 50\\\midrule
Track IoU / hold / confirmation &0.25 / 3.0 s / 2 hits\\
Immediate danger bottom / area &0.65 / 0.02 of normalized image\\
Motion history / minimum &24 samples, 1.5 s / 5 samples, 0.4 s\\
Maximum motion gap / deadband &0.75 s / $\pm0.04$ s$^{-1}$\\
Input freshness / advisory lifetime &0.8 s / 3.5 s\\
Perception / advisor publishing &0.5 Hz / 5 Hz (republish, not new frames)\\
Center / side risk thresholds &0.48 / 0.58\\\bottomrule
\end{tabularx}
\end{table}

\begin{table}[H]\centering\small
\caption{Provisional robotics parameters. Metric-prototype parameters are not active camera-simulation evidence.}
\begin{tabularx}{\linewidth}{p{.38\linewidth}Y}\toprule
Setting & Value / scope\\\midrule
Mass / equilibrium height &12.4 kg / 0.177 m; uncalibrated model\\
Heave stiffness / damping &650 / 90; simulator coefficient units\\
Roll-pitch restoring / damping &90 / 24; simulator coefficient units\\
Surge linear / quadratic drag &4 / 20\\
Sway linear / quadratic drag &20 / 80\\
Yaw linear / quadratic drag &4 / 25\\
Active cruise / target speed cap &0.15 / 0.30 m/s\\
Active normalized forward / steering cap &0.08 / 0.04\\
Controller rate / slew limit &10 Hz / 0.15 s$^{-1}$\\
Fan conversion / adapter watchdog &75 N per normalized unit / 0.5 s wall time\\\midrule
Experimental metric grid &100 by 100 cells, 0.2 m; unknown by default\\
Free / occupied evidence lifetime &0.8 / 2.0 s\\
Swept-footprint horizon / integration &3.0 s / 0.2 s\\
Prototype speed / yaw-rate / acceleration &0.4 m/s / 0.35 rad/s / 0.2 m/s$^2$\\
Hull radius / margin / drift allowance &0.55 m / 0.25 m / 0.15 m/s\\\bottomrule
\end{tabularx}
\end{table}
\FloatBarrier
\bibliographystyle{plainnat}
{\raggedright\bibliography{references}}
\end{document}